# A Survey on Resilient Machine Learning

Atul Kumar, Sameep Mehta
IBM Research, India

## ABSTRACT

Machine learning based system are increasingly being used for sensitive tasks such as security surveillance, guiding autonomous vehicle, taking investment decisions, detecting and blocking network intrusion and malware etc. However, recent research has shown that machine learning models are venerable to attacks by adversaries at all phases of machine learning (e.g., training data collection, training, operation). All model classes of machine learning systems can be misled by providing carefully crafted inputs making them wrongly classify inputs. Maliciously created input samples can affect the learning process of a ML system by either slowing the learning process, or affecting the performance of the learned model or causing the system make error only in attacker's planned scenario. Because of these developments, understanding security of machine learning algorithms and systems is emerging as an important research area among computer security and machine learning researchers and practitioners. We present a survey of this emerging area.

## Categories and Subject Descriptors

D.4.6 [**Security and Protection**]: Invasive software; I.2.6 [**Learning**]: Concept learning; I.5.1 [**Models**]: Neural nets; I.5.2 [**Design Methodology**]: Classifier design and evaluation.

## General Terms

Algorithms, Design, Security, Theory.

## Keywords

Resilience, Adversarial Learning, Computer Security, Intrusion Detection.

## 1. INTRODUCTION

Over last few years, machine leaning has become a prominent technological tool in several application areas such as computer vision, speech recognition, natural language understanding, recommender systems, information retrieval, computer gaming, medical diagnosis, market analysis etc. In many areas, it is no longer a promising but immature technology as machine learning based systems have reached close to human level performance. Most of machine learning techniques build models using example data (training data). These models along with algorithms can be used to make predictions on data not seen before.

Learning and building models using training data provides hackers opportunities to attack machine learning algorithms by playing with the features and decision boundaries of the model. An adversary can craft malicious inputs to attack the performance or efficiency of a machine learning algorithm. Some systems where data distribution is not fully known at training time, use data examples in future to continuously train the system to adjust for the changing data distribution. An adversary can contribute malicious inputs to 'poison' the system. Others can dupe an already trained system by creating input data that exploits the system into making glaring errors.

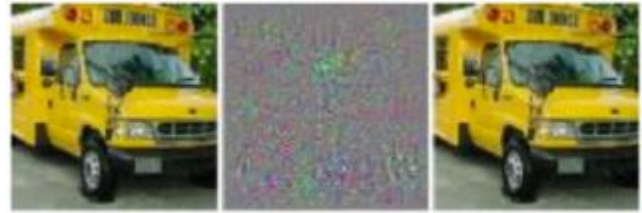

**Figure 1: Creating adversarial example using noise**
**(Image credit: Szegedy et al. [1])**

For example, researchers have demonstrated [1], how to fool an image classification system by making tiny changes to the input images. Figure 1 shows three images. On the left is an image that the system correctly classifies as a school bus. Image in center is a noise which when added to the left image creates an image shown on the right which still looks like a school bus to a human observer. But the system now classifies this image (right) as an ostrich. These techniques can be used by hackers to evade the system in making it accept malicious content as a genuine one. With machine learning becoming an important tool in strategically important applications such as security surveillance and background check for visa decisions etc., it is important to understand these attacks and make machine learning algorithms more robust against these attacks.

If a hacker does not already know the algorithm, he first tries to learn the algorithm and its underlying model (e.g., logistic regression, neural network, decision trees etc.). Sometime, the hacker may only be interested in learning the model so that he can build his own 'copy' of the system using the learned model. This may be useful if the application is offered as a service via APIs and users are charged per use of these APIs. A hacker can create a sequence of inputs and then by observing outputs of the system corresponding these inputs, he can build a local model that may be very close to the model used by the original system. Depending on the pricing and the license terms of the API usage, a hacker may be able to 'steal' the model using very small amount of money. Tramer et al demonstrated at USENIX Security Symposium 2016 [2] that models can be extracted from popular online machine learning services such as BigML and Amazon Machine Learning with a relatively small number of API calls.

Another category of attacks on machine learning systems is to provide adversarial input during the training phase and compromise the learning by affecting its efficiency or introducing some bias. Many systems allow users to provide training data samples for online training of the system. Collecting training data from people spared across geographies is immensely valuable in many applications to have good data distribution. But opening the system to public for providing input data also opens a system to malicious input created by hackers to 'poison' the system. Microsoft's twitter chatbot Tay started tweeting racist and sexist tweets in less than 24 hours after it was opened to public for learning [3].

This survey categorizes major works in Adversarial Machine Learning area in three broad categories. First set consists of research focusing on learning algorithms and its models by providing carefully crafted inputs and then observing the output to build local copies of the models. Second set consists of techniques focusing on evasion attacks. And the third set combines the work focusing on poisoning attacks. These are not disjoint sets and many of the works overlaps across these categories.

Rest of this paper is organized as follows. In Section 2, we discuss some earlier work on creating adversarial input to attack classification systems such as spam filters and anti-virus/malware systems. Section 3 discusses exploratory attacks that aim to learn algorithms and models of machine learning systems under attack. Section 4 discusses work related to evasion attacks. Section 5 discusses work covering poisoning attacks. Section 6 provides a summary in a table. Section 7 concludes the survey with a discussion on trends and research directions in this area. Appendix-A lists libraries and other open source software/data-repositories useful for Adversarial Machine Learning research.

## 2. EARLIER RELATED WORK

Some early work in attacking learning algorithms with malicious input come from anti-spam filters, anti-malware and biometric verification domains. A classifier is designed that works on input samples. It automatically determines whether a sample falls into a malicious target class (e.g., spam email or malware/worm) or a safe/genuine class. The classifier is typically generated automatically using learning by analyzing labeled training samples.

Data mining algorithms normally assume that data gathering activities are independent of data mining algorithms. However, an adversary manipulates data actively such that many false negatives are produced.

Dalvi et al [4] proposed a game theory based approach to design classifiers. The classification process is viewed as a game between classifier and adversary. An optimal classifier is produced for adversary's optimal strategy. The classifier thus automatically adjusts to adversary's evolving inputs. Their experiments with spam filtering show that such classifiers outperform standard classifiers. Bruckner and Scheffer [5, 6, 7] proposed single-shot prediction games in which the cost functions of classifier and adversary are not necessarily completely opposed to each other. They identified conditions such that a prediction game achieves a unique Nash equilibrium. They proposed techniques to develop algorithms that find the equilibrium prediction models. These techniques work well in email spam filtering.

Often, it is assumed that adversaries have good knowledge of classifiers which may be unrealistic. Lowd and Meek [8] defined an adversarial classifier reverse engineering (ACRE) learning problem. Without any knowledge of a classifier, the task is to learn enough about it to construct adversarial attacks. Their algorithms targeted liner classifiers with either continuous or Boolean features and used spam filter data to demonstrate effectiveness.

Barreno et al [9] provided a taxonomy of different types of attacks on machine learning algorithms and an analytical model giving a lower bound on attacker's work function. Proposed taxonomy puts attack models in three major categories – *influence*, *specificity* and *security violation*. Influence attacks are further divided into two sub-categories – *causative* attacks change the training process by having some control over the training data; and *exploratory* attacks try to discover information using some probes. Under *specificity* attacks, *targeted* attacks are those that only focus their attack on a specific point or a small set of points; whereas *indiscriminate* attacks are those where attackers are flexible and involve some general class of points (e.g., any false negative). *Security violation* attacks include *integrity* attacks where intrusion points are made to be classified as normal; *availability* attacks are broader than integrity attack. They aim to cause many classification errors (both false negatives and false positives) so that system effectively becomes unusable.

In anti-malware/virus software and network security domains, signature detection approaches are widely used. To evade signature-based intrusion detection systems, attackers employ polymorphic techniques to generate attack instances that do not share fixed signatures. Anomaly detection is used to guard against such attacks because even though attackers can use polymorphic techniques to make attack instances look different from each other, they cannot make them look normal. Fogla et el [10] proposed a new class of polymorphic attacks, called polymorphic blending attacks (a sub class of mimicry attacks). These attacks can evade byte-frequency based network anomaly intrusion detection systems by matching the statistics of mutated attack samples with normal samples. Newsome et al [11] designed practical attacks against learning and used them on automatic polymorphic worm signature generation algorithms effectively. In their approach, an adversary builds labeled samples. Training with these samples prevent or severely delay generation of good classifiers. They show that a delusive adversary can obstruct learning whose samples are all otherwise correctly labeled.

In view of attacks on learning based classifiers, it is desirable to make these classifiers robust against such attacks. One obvious thing is to not assign too much weight to a single feature to increase robustness of a classifier. Regularization is used to spread the weight more evenly between the features. However, regularization is a very generic technique and may not be suitable to specific classification tasks. Globerson and Roweis [12] introduced an algorithm to avoid single feature over-weighting by analyzing robustness using a game theoretic formalization. These classifiers are optimally resilient to deletion of features in a minimax sense. They constructed such classifiers using quadratic programming. These classifiers were tested in spam filtering and handwritten digit recognition tasks. Kolcz and Teo [13] introduced a new method to find the lower bound of classifier robustness. Simple averaged classifiers can improve robustness considerably. They also proposed a feature reweighting algorithm to improve robustness and performance of classifiers. Biggio et al [14] experimentally investigate whether the technique proposed in [13] can be implemented using bagging [15] and random subspace method (RSM) [16] – two well-known techniques for multiple classifier construction.

In applications such as biometric verification and authentication, characteristics used to meet some basic requirements such as universality, distinctiveness, permanence, etc. But in practice, no biometric trait fully meets these requisites which means no single biometric mode is error free. Multiple biometric modalities are used to minimize these errors. Rodrigues et al [17] proposed two fusion schemes that aim to increase the robustness of multimodal biometric systems. First is a likelihood ratio based fusion scheme and the other is based on fuzzy logic. In addition to matching score and sample quality score, the proposed fusion schemes also considers intrinsic security of different biometric system being used. They demonstrate that these methods are more robust against spoof attacks than traditional fusion methods.

## 3. EXPLORATORY ATTACKS

Exploratory attacks do not attempt to influence training; instead they try to discover information from the learner that includes discovering which machine learning algorithm/technique is being used by the system, state of the underlying model and training data.

### 3.1 Model Inversion

Fredrikson et al introduced the term *model inversion* in [18]. In pharmacogenetics, machine learning techniques are used to assist in medical treatments based on patient's genotype and other background. Maintaining privacy about patients' personal and medical records is an important requirement in healthcare domain and mandated by law in many nations. Fredrikson et al showed that by using the model (black box access) and some demographic information about a patient, an attacker can predict the patient's genetic markers. This attack works in a setting in which the sensitive feature being inferred is drawn from a small set. Differential privacy is an often used as a solution for situations like this. Authors showed that differential privacy when used with appropriate privacy budgets can prevent their model inversion attacks but it may impact the clinical efficacy therefore putting patients to some risk. It is not known whether *model inversion* attacks propose in [18] work outside their settings. Fredrikson, Jha and Ristenpart extended their previous work to develop a new class of model inversion attack [19]. This new model inversion attack uses the confidence percentage provided with predictions. They tested their new attack on commercial ML-as-a-service APIs. The attack infers sensitive features used as inputs to decision tree models for lifestyle surveys, as well as to recover images from API access to facial recognition services. Figure 2 shows a recognizable image of a person produced by an attacker using *new model inversion attack*. Only API access to a facial recognition system and the name of the person whose face is recognized by it were available to the attacker. In another experiment, they attacked a decision trees for lifestyle surveys and could estimate whether a respondent in the survey admitted to cheating on their significant other. The paper discusses some countermeasures to *model inversion* attack and show that systems can be secured against these kinds of attacks with negligible degradation to utility.

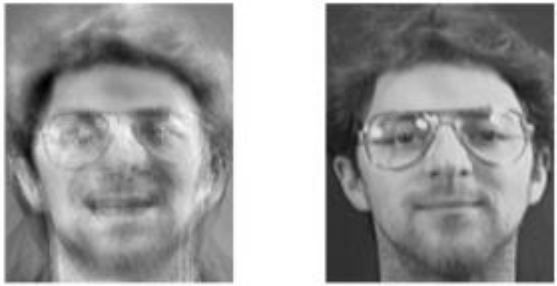

**Figure 2: An image recovered using a new *model inversion* attack (left) and the actual training set image (right).**

**(Image Credit: Fredrikson et al [19])**

### 3.2 Inferring useful information

Even though the major machine learning algorithms are publicly known, the training data used to build a proprietary model may not be publically available and may be protected as trade secrets. Some machine learning classifiers can unconsciously revel the statistical information. Ateniese et al [20] showed that it is possible to infer unexpected but useful information from machine learning classifiers. They build a meta-classifier which is trained to hack other classifiers, obtaining meaningful information about their training sets. Using the methodology proposed, an adversary infers statistical properties from the relationship among dataset entries and not the attributes of the dataset. They experimented with a speech recognition classifier that uses Hidden Markov Models and extracted information such as the accent of the speakers that is not supposed to be captured explicitly by the model and which is not an attribute of the training set. In another case study, they showed that it is possible to determine whether a certain type of network traffic was included in the training set of an Internet traffic classifier trained on network data flow data.

Another type of inference attack is membership inference attack. Given the black box access to a model (e.g., via public APIs) and a data record, an attacker may be interested in knowing whether that data record was part of the training set of the model. Shokri et al [21] used membership attacks on classification models trained by commercial "ML as a service" providers such as Google and Amazon. They trained their own inference model and then observed the differences in predictions by their own model and by the target model. They compared differences for the inputs that were used to train their own model versus the inputs that were not. Some realistic datasets and classification tasks such as a hospital discharge dataset were used for experiments. They demonstrated that these models are vulnerable to membership inference attacks.

### 3.3 Model Extraction using Online APIs

Machine learning as a service for applications such as predictive analytics are deployed with publicly accessible query interfaces (APIs). These models are deemed confidential due to their sensitive training data, commercial value, or other reasons such as use in security applications. Access is provided on a pay-per-query basis. In such situations, an adversary has black-box access but no prior knowledge of the machine learning model's parameters or training data.

Tramer et al [2] presented simple attacks to extract target machine learning models for popular model classes such as logistic regression, neural networks, and decision trees. Model extraction attacks were demonstrated on popular online ML-as-a-service providers such as BigML and Amazon Machine Learning. Their attacks were complete black box and the adversary does not even need to know the model type or any distribution information about training data. They could build local models that are functionally very close to the target. In some experiments, their attacks extracted the exact parameters of the target (e.g., the coefficients of a linear classifier or the paths of a decision tree). In situations where the model type, parameters or features of the target were not known, they used an additional preliminary attack step to reverse-engineer these model characteristics. Machine learning prediction APIs of major online services such as Google, Amazon, Microsoft, and BigML all return precision confidence values along with class labels. Moreover, they work with partial queries lacking one or more features. These features can be exploited for model extraction attacks. The confidence value for logistic regression is a simple log-linear function $1/(1+e^{-(w \cdot x + \beta)})$ of the d-dimensional input vector $\mathbf{x}$. Therefore, an attacker may solve w and $\beta$ that define the model by querying d+1 random d-dimensional inputs for the unknown d+1 parameters. Such equation-solving attacks extend to multiclass logistic regressions and neural networks, but do not work for decision trees. For decision trees, a confidence value implies the number of training data samples labeled correctly on an input's path in the tree. However, these confidence values can still be used as pseudo-identifiers for paths in the tree therefore assisting in discovering tree's structure. Omitting confidence values from outputs is an obvious countermeasure against these attacks. Authors proposed new attacks inspired by an agnostic learning algorithm [22]. Their new attacks extract models from more than 99% of targets for a variety of model classes but need up to 100 times more queries than equation-solving attacks.

Papernot et al [23] introduced a practical black-box attack on remotely hosted deep neural networks (DNN) with no knowledge of either the model internals or their training data. An adversary observes output label given by the DNN to chosen inputs. They train a local model using inputs synthetically generated and labeled by the target DNN. The trained local model can be used for other attacks such as crafting adversarial examples for mounting evasion attacks on the target DNN.

## 4. EVASION ATTACKS

Evasion attacks are the most prevalent type of attack on a machine learning system. Malicious inputs are carefully crafted to evade detection which essentially means that input is modified to make the machine learning algorithm classify it as a safe one instead of malicious.

### 4.1 Adversarial Examples

Szegedy et al [1] found that deep neural networks (DNN) learn input-output mappings that are fairly discontinuous. One can cause a DNN to wrongly classify an image by applying a specifically crafted modification (found by maximizing the network's prediction error) that is difficult to distinguish by a human viewer. The same change to the image can cause a different network, trained on a different subset of the dataset, to incorrectly classify the same image. This property of deep neural network can be exploited to create any number of adversarial inputs from the normal inputs.

Practical Black-Box Attacks method proposed by Papernot et al [23] misclassified 84.24% of the crafted adversarial examples on MetaMind (an online deep learning API) DNN. They also used logistic regression substitutes to craft adversarial examples for Amazon and Google ML APIs and found misclassification rate of 96.19% and 88.94% respectively.

Papernot et al [24] show that adversarial attacks are also effective when targeting neural network policies in reinforcement learning. Adversaries capable of introducing small perturbations to the raw input can significantly degrade test-time performance. The strategy is to train a local substitute DNN using a synthesized data set. Input data is synthesized but the label assigned is what the target DNN assigns to it and observed by the adversary. Adversarial examples are generated by using the substitute parameters known to adversary. These are misclassified by both target DNN and the substitute DNN created locally because they both have the same decision boundaries. To create a small perturbation so that the changed image looks similar to the original one, an algorithm named *fast gradient sign method* [25]. The cost gradient is computed for pixels and the target pixels (areas) for perturbation is identified. Another algorithm by Papernot et al. [26] can cause a misclassification for samples from any legitimate source class to any chosen target class. That is, any image can be changed slightly such that it is classified to a desired class (say ostrich) by the DNN. Therefore, a school bus image can be changed in such a way that to humans, it still looks like a bus but the DNN recognizes it as an ostrich (for that matter any class chosen by the adversary). Input components are added to a perturbation in order of decreasing adversarial saliency value until the resulting adversarial sample is misclassified by the mode.

Goodfellow et al [25] argue that the primary cause of neural networks' vulnerability to adversarial perturbation is their linearnature. The most intriguing fact about neural networks is their generalization across architectures and training sets. Using this view, they discussed a simple and fast method of generating adversarial examples. This approach can be used to generate examples for adversarial training to reduce the test set error.

Papernot et al [26] introduced a class of algorithms to create adversarial inputs based on a precise understanding of the mapping between inputs and outputs of DNNs. They defined a hardness measure to evaluate the vulnerability of different sample classes to adversarial perturbations. They defined a predictive measure of distance between a benign input and a target classification to describe preliminary defenses against adversarial samples.

Moosavi-Dezfooli et al [27] proposed the DeepFool algorithm to efficiently compute perturbations that fool deep networks thus quantifying the robustness of these classifiers. This computation of robustness can be used to make classifiers more robust.

### 4.2 Generative Adversarial Networks (GANs)

Goodfellow et al [28] introduced Generative adversarial networks. They are implemented by simultaneously training two models: a generative model $G$ that captures the data distribution, and a discriminative model $D$ that estimates the probability that a sample came from the training data rather than $G$. The training procedure for $G$ is to maximize the probability of $D$ making a mistake. This can be viewed as a competition between a team of counterfeiters and a team of police. If generative model is assumed to be producing fake currency such that it can pass without detection, then the discriminative model is trying to detect the counterfeit currency. Competition leads both teams to improve their methods until the counterfeits cannot be distinguished from the genuine currency.

Kos et al [29] presented three classes of attacks on the VAE and VAE-GAN architectures [30]. The first attack leverages classification-based adversaries by attaching a classifier to the trained encoder of the target generative model. The second attack directly uses the VAE loss function to generate a target reconstruction image from the adversarial example. And the third attack directly optimizes against differences in source and target latent representations.

### 4.3 Query Strategies for Evasion

An adversary systematically creates queries for a classifier to elicit information that allows the attacker to evade detection. Nelson el al [31] present query algorithms to construct undetected instances for convex-inducing classifiers. Only polynomially-many queries in the dimension of the space and in the level of approximation are required with approximately minimal cost. The family of convex-inducing classifiers partition their feature space into two sets, one of which is convex. This family is a super set of the family of linear classifiers. They demonstrated that near optimal evasion can be achieved for convex-inducing classifiers without a need to know the classifier's decision boundary.

### 4.4 Adversarial Classification

In traditional classification, an input is classified as one of classes. In an adversarial setting, an adversary can manipulate input instances to avoid being so classified. Learning to distinguish good inputs from malicious ones is known as adversarial classification. Vorobeychik and Li [32] presented a general theoretical analysis of the problem of adversarial classification. They generalized adversarial classifier reverse engineering (ACRE) process to demonstrate that if a classifier can be efficiently learned, it can also be efficiently reverse engineered. This result is extended to randomized classification schemes showing that effectiveness of reverse engineering depends on the defender's randomization scheme. They characterized optimal randomization schemes in presence of adversarial reverse engineering and classifier manipulation. They observed that the defender's optimal policy tends to be either to randomize uniformly (ignoring baseline classification accuracy) or not to randomize at all (i.e, targeted attacks or indiscriminate attacks).

Adversaries are not static data generators, but make a deliberate effort to evade the classifiers deployed to detect them. Li and Vorobeychik [33] studied the problems of modeling the objectives of such adversaries and the algorithmic problem of accounting for rational, objective-driven adversaries.

## 4.5 Evasion Attacks on Text Based Systems

Perturbation techniques for image or audio based system cannot directly work on text based systems. That is because an important requirement of the perturbation used is to change the image or audio such that it still looks good to a human observer/listener. Whereas in a text, changing words by adding/deleting characters or changing sentences by adding/deleting words may make the sentence/word meaningless or change its meaning significantly and therefore cannot remain unnoticed by a human reader. Therefore, a perturbation technique must change the text such that it still looks good/suspicious to a human observer but machine learning system fails to classify it correctly after perturbation. For example, a spam email carrying an advertisement should still carry the advertisement message but fool the spam filtering system in classifying it as a regular email.

Creating adversarial inputs for text classification systems seems to be a harder problem than doing the same for the image or audio classification. Some recent work has shown that it is possible to systematically create such adversarial inputs. Liang et al [45] discuss the problem of creating perturbation. They propose three techniques named *insertion*, *modification*, and *removal*, to generate adversarial samples for given text. They compute cost gradients (originally proposed in [25] for images and proven to be effective in [26] and [27]) to decide what and where should be inserted, what and how to modify and what should be removed from a text sample. However, using the fast gradient sign method (FGSM) of [25] directly makes the text unreadable. Using cost gradient, they identify the text items that possess significant contribution to the classification. Then instead of changing the characters arbitrarily, they use one or more of insertion, modification and removal to craft an adversarial sample for a given text.

They demonstrate using experiments that standard text classification systems can be deceived into misclassifying input text samples as any desirable classes by using their perturbation techniques without compromising the utility of the input text. They also show that deep neural network based text classifiers are also prone to such attacks. They have shown that by adding just one word at a specific place in a paragraph, or by misspelling just one instance of a word in a paragraph, the system can be fooled in classifying a text paragraph incorrectly with very high confidence. For example, a paragraph describing 1939 film *Maisie* is correctly classified by the system as about films with 99.6% confidence. They slightly misspelled a word ("film" to "flim") at a particular place in the paragraph and the system classifies the modified paragraph as about companies even though the paragraph still contains other correctly spelled instances of the world "film".

## 4.6 Evaluating Classifiers Security Against Evasion Attacks

Adversarial scenario is often not considered at design time. A framework to evaluate potential performance degradation under potential attacks is proposed by Biggio et al [34]. The proposed framework evaluates classifier security empirically. It formalizes and generalizes the main ideas proposed in pattern classification theory and design methods. This framework is used to build a gradient-based approach [35] to assess the security of widely-used classification algorithms against evasion attacks. This provides the designer a better picture of the classifier performance under potential evasion attacks. The designer therefore can perform a more informed model selection (or parameter setting).

## 5. POISONING ATTACKS

In poisoning attacks, attackers try to influence training data to influence the learning outcome. The purpose of poisoning attacks may vary from affecting the performance of learning algorithm to deliberately introducing specific biases in the model. In many applications, training is not a one-time job and model is often retrained to accommodate for the change in data distribution. In some situation, data collection is crowdsourced and many users provide data sample that are used to continuously train the model. Some domains such as network intrusion detection, spam filtering, malware detection etc. are highly suspect of poisoning attacks but any machine learning system can be a victim of poisoning attacks.

## 5.1 Network Intrusion Detection

For intrusion detection, statistical machine learning based techniques build a model for normal behavior from training data and then detect attacks that deviates from that model. Adversaries try to manipulate the training data so that the learned model treats intrusion also as normal network behavior.

Rubinstein et al [36] show how attackers can substantially increase their chance of successfully evading detection by only adding moderate amounts of poisoned data. Such poisoning disturbs the balance between false positives and false negatives resulting in dramatically reduced effectiveness of the system. They proposed a robust PCA-based detector called 'antidote'. It is based on techniques from robust statistics. They show that poisoning has little effect on the robust model.

Kloft and Laskov [37] analyzed the performance of online centroid anomaly detection in an adversarial setup. They derived bounds on the effectiveness of a poisoning attack against centroid anomaly detection under different conditions. While poisoning attacks can be successful in the unconstrained case, they become arbitrarily difficult if external constraints are properly used. They used real traces of HTTP and exploit traffic and confirmed the tightness of proposed theoretical bounds.

## 5.2 Poisoning Support Vector Machines

Xiao and Eckert [38] address the problem of label flips attack. In this attack, an adversary poisons the training set by flipping labels. An optimization framework is formulated to finds label flips that maximize the classification error. They proposed an algorithm for attacking support vector machines (SVMs).

Biggio et al [39] discuss a family of poisoning attacks against Support Vector Machines (SVM). These attacks introduce specially crafted training data to increases SVM's test error. Learning algorithms often assume that training data comes from a well-behaved distribution which is not true in an adversarial setting. They demonstrated that an intelligent adversary can predict the change of the SVM's decision function to some extent by using malicious input. The adversary then uses this ability to craft malicious samples. Proposed attack uses a gradient ascent strategy where the gradient is computed based on properties of the SVM's optimal solution. The aim of such attacks is to increase classifier's error rate.

## 5.3 Factorization-Based Collaborative Filtering

Li et al [40] introduced a data poisoning attack on collaborative filtering systems. They demonstrated how a powerful attacker with full knowledge of the learner can generate malicious data to maximize his objectives while mimicking normal user behavior to avoid detection. The assumption about complete knowledge is extreme but it enables a robust assessment of the vulnerability of collaborative filtering schemes to highly motivated attacks. Authors considered two popular factorization-based collaborative filtering algorithms - the alternative minimization formulation and the nuclear norm minimization method.

## 5.4 Defensive Distillation

Papernot at al [41] introduced a defensive mechanism called *defensive distillation* that reduces the effectiveness of adversarial samples on deep neural networks (DNNs). Distillation is a training procedure that was designed to train a DNN using knowledge transferred from a different DNN [46][47]. The motivation behind the knowledge transfer is to reduce the computational complexity of DNN architectures by transferring knowledge from larger architectures to smaller ones. This facilitates the deployment of deep learning in resource constrained devices that cannot rely on powerful GPUs to perform computations. A new variant of distillation is proposed for defense training. Instead of transferring knowledge between different architectures, knowledge extracted from a DNN is used to improve its own resilience to adversarial samples. An analytical investigation is presented for the generalizability and robustness properties granted by defensive distillation when training DNNs. Two DNNs were placed in adversarial settings to empirically study the effectiveness of defensive distillation. They show that defensive distillation can reduce effectiveness of sample creation from 95% to less than 0.5% on the DNNs used in their study. This can be explained by the fact that distillation reduces by a factor of 1030 the gradients used in adversarial sample creation. Distillation also increases by 800% the average minimum number of features required to be modified for creating adversarial samples on one of the DNNs used in their experiments.

## 5.5 Semi-Supervised Text Classification

Adversarial training is used for regularizing supervised learning algorithms and virtual adversarial training extends supervised learning algorithms to the semi-supervised setting. Both methods require making small perturbations to several entries of the input vector. This is inappropriate for sparse high-dimensional inputs such as one-hot word representations. Miyato et al [42] extend adversarial and virtual adversarial training to the text domain. by Perturbations is applied to the word embeddings in a recurrent neural network and not to the original input. This method achieves good results on multiple benchmark for semi-supervised and supervised tasks. Analysis shows that the learned word embeddings improved in quality and that the model is less prone to overfitting while training.

## 6. SUMMARY

**Early Work**

**Applications:** Spam filters, anti-virus/ malware, network intrusion detection, biometric verification and authentication

**Approaches**:

*Game-theory based approaches* – game between classifier and adversary. optimal classifier to automatically adjusts to adversary's evolving inputs; find the equilibrial prediction models

*Signature-based intrusion detection systems* - polymorphic techniques to generate attack instances that do not share fixed signatures - attackers can use polymorphic techniques to make attack instances look different from each other.

*Polymorphic blending attacks* - can evade byte-frequency based network anomaly intrusion detection systems by matching the statistics of mutated attack samples with normal samples.

*Making classifiers robust* - not assign too much weight to a single feature; game theoretic formalization to avoid over weighting single feature.

*Multimodal biometric systems* - likelihood ratio based fusion scheme and fuzzy logic based fusion scheme

**Attacks**: *exploratory, evasion* and *poisoning*

**Exploratory attacks**:

*Model Inversion*: black box access to model and some demographic information about a person, an attacker can predict private information such as genetic markers from a healthcare system.

*Inferring information*: an adversary can infer statistical properties from the relationship among dataset entries

*Membership inference attack*: given the black box access to model and a data sample, it can be inferred whether that data record was part of the training set or not.

*Model Extraction using Online APIs*: local models can be built that function very similar to proprietary models for which only API access is available. Confidence score and partial values for API are used to find key coefficients of the model.

**Evasion attacks**:

*Adversarial Examples*: systematic adversarial perturbation; DeepFool algorithm to efficiently compute perturbations that fool deep networks

*Generative Adversarial Networks (GANs)*: simultaneously training two models: a generative model $G$ that captures the data distribution, and a discriminative model $D$ that estimates the probability that a sample came from the training data rather than $G$.

*Adversarial classification*: Learning to distinguish good inputs from malicious ones is known as adversarial classification. Useful in adversarial training.

*Text-based systems*: text classification systems can be fooled by carefully inserting, modifying or removing some text such that the meaning of text does not change for a human user.

**Poisoning attacks:**

*Network Intrusion Detection*: input samples to disturb the balance between false positives and false negatives therefore reducing effectiveness.

*Support Vector Machine Poisoning*: label flips attack; adversary can predict the change of the SVM's decision function to some extent by using malicious input. This can be used to craft malicious samples.

*Defensive Distillation*: Distillation is a training procedure that was designed to train a DNN using knowledge transferred from a different DNN. This technique is used for defense training.

# 7. EMERGING RESEARCH DIRECTIONS

Generating adversarial examples to fool machine learning algorithms in making incorrect classification and making machine learning systems robust against these inputs are active research areas. But it is fundamentally hard problem to defend against adversarial examples because it is hard to build a theoretical model for crafting adversarial inputs. Adversarial inputs are solutions to an optimization problem that is non-linear and non-convex for many machine learning models, including neural networks. Since good theoretical tools for describing the solutions to these optimization problems do not exist, it is very hard to put forward a theoretical argument that a defense strategy would rule out a set of adversarial inputs. Like other computer security areas such as computer viruses/malwares, a defense technique makes a system robust against one type of attack. When a new vulnerability is discovered by an attacker, a new defense is required to build for that attack. Designing an adaptive defense against an adaptive attacker is an important research area.

Using defensive distillation to guard a system against adversarial examples is an interesting research direction. A neural network is used to label images with probability vectors instead of single labels. A new neural network is then trained using the probability vector labels. This makes the second neural network less prone to over-fitting. Papernot and his fellow researchers have shown that to fool such networks, eight times more distortion to the image is needed compared to what is needed without distillation. There seems to be scope further improve defensive distillation methods to make networks more robust.

Building benchmarks to measure a machine learning model's performance performs against an adversary is another open problem. Ideally, there should be standard set of benchmarks for measuring accuracy of an ML algorithm and there should be standard benchmarks for measuring its performance in adversarial settings.

If we see the history of software development, security was added to software products at a very late stage, often after a functional product is built and tested for its functional and performance related requirements. Only recently, security became an important consideration at requirement analysis and design stage. Machine learning systems can be considered in their early days where security is considered only after the system is attacked or some vulnerability is discovered. Security in machine learning systems should be built from the start and not as an afterthought. Some research is due on evolving software engineering methodology for building machine learning system.

Another emerging research area in adversarial ML domain is of Generative Adversarial Networks. Theoretically, generative model and discriminative model in a GAN should be best at the Nash equilibrium. But a gradient descent is guaranteed to get to the Nash equilibrium only in the convex case. And for other ML models, it is not even possible to reach equilibrium. If players are represented by neural networks then they can keep adapting forever and would never converge therefore never reaching the equilibrium. Fixing the non-convergence problem in GANs is an open problem as of now.

Finally, some fundamental work is due to change the learning process of machines. Quoting a story from Dave Gershgorn's article in Popular Science [44]:

> *In the early 1900s, Wilhelm von Osten, a German horse trainer and mathematician, told the world that his horse could do math. He would ask his horse, Clever Hans, to compute simple equations. In response, Hans would tap his hoof for the correct answer. Two plus two? Four taps.*
>
> *Psychologist Carl Stumpf found that Clever Hans wasn't solving equations, but responding to visual cues. Hans would tap up to the correct number, which was usually when his trainer and the crowd broke out in cheers. And then he would stop. When he couldn't see those expressions, he kept tapping and tapping.*

A lot of today's machine learning and artificial intelligence work like Hans. We know how to build systems that can learn enough to give correct answers but without really understanding the information. That makes them easy to deceive. It is like guessing and constructing a digital circuit with $n$ inputs by looking at $k$ random items in its truth table where $k << 2^n$.

# Appendix-A: Open Source Software/Libraries

## cleverhans

Named after the early 20[th] century house Clver Hans who his trainer claimed could do math, cleverhans is a software library for benchmarking vulnerability to adversarial examples. The library is maintained by Ian Goodfellow and Nicolas Papernot, two leading researchers in adversarial machine learning.

Repository URL: https://github.com/openai/cleverhans

This repository contains the source code for cleverhans, a Python library to benchmark machine learning systems' vulnerability to adversarial examples.

## textfool

Plausible looking adversarial examples for text classification. It provides a function that "paraphrases" a text by replacing some words with their WordNet synonyms, sorting by GloVe similarity between the synonym and the original context window. Relies on SpaCy and NLTK.

Repository URL: https://github.com/bogdan-kulynych/textfool

## AdversariaLib and ALFASVMLib

AdversariaLib is an open-source python library for the security evaluation of machine learning (ML)-based classifiers under adversarial attacks.

ALFASVMLib is an open-source Matlab library that implements a set of heuristic attacks against Support Vector Machines (SVMs). The goal of such attacks is to maximally compromise the SVM's classification accuracy by mislabeling a given fraction of training samples. They are indeed referred to as adversarial label flip attacks

These libraries are maintied by PRA Lab, University of Cagliari, Italy.

URLs: http://pralab.diee.unica.it/en/AdversariaLib and http://pralab.diee.unica.it/en/ALFASVMLib

## deep-pwning: Metasploit for machine learning

Deep-pwning is a lightweight framework for experimenting with machine learning models with the goal of evaluating their robustness against a motivated adversary. This framework is built on top of Tensorflow.

URL: https://github.com/cchio/deep-pwning